\documentclass[letterpaper]{article} %
\usepackage[draft]{aaai2026}  %
\usepackage[table]{xcolor}
\usepackage{times}  %
\usepackage{helvet}  %
\usepackage{courier}  %
\usepackage[hyphens]{url}  %
\usepackage{graphicx} %
\usepackage{natbib}  %
\usepackage{caption} %

\usepackage{algorithm}
\usepackage{algorithmic}

\usepackage{newfloat}
\usepackage{listings}
\DeclareCaptionStyle{ruled}{labelfont=normalfont,labelsep=colon,strut=off} %
\floatstyle{ruled}
\newfloat{listing}{tb}{lst}{}
\floatname{listing}{Listing}
\newcommand{\myparagraph}[1]{\textbf{#1}}
\newcommand{\myparagraphnoindent}[1]{\noindent{\textbf{#1}}}
\newcommand{\designer}{Designer}
\newcommand{\coder}{Coder}
\newcommand{\visualInspector}{Visual Inspector}
\newcommand{\method}{3DFroMLLM}

\usepackage{multirow}
\usepackage{dingbat}
\usepackage{tikz}
\usepackage{adjustbox}
\usepackage{longtable, array, booktabs}
\usepackage{savesym}
\savesymbol{checkmark}
\usepackage{amssymb}
\usepackage{amsmath} %
\usepackage{dcolumn}
\usepackage{siunitx}
\usepackage[percent]{overpic}
\usepackage{hyperref} %
\usepackage{cleveref}

\title{3DFroMLLM: 3D Prototype Generation only from\\ Pretrained Multimodal LLMs}
\author{
    Noor Ahmed\textsuperscript{\rm 1,\thanks{Corresponding author mail: \textrm{noor.ahmed@utn.de}}}, 
    Cameron Braunstein\textsuperscript{\rm 2},
    Steffen Eger\textsuperscript{\rm 1}, 
    Eddy Ilg\textsuperscript{\rm 1}
}
\affiliations{
    \textsuperscript{\rm 1}University of Technology Nuremberg \\
    \textsuperscript{\rm 2}Saarland University

}

\usepackage{bibentry}

\begin{document}

\maketitle

\begin{abstract}
Recent Multi-Modal Large Language Models (MLLMs) have demonstrated strong capabilities in learning joint representations from text and images. However, their spatial reasoning remains limited. 
We introduce \method{}, a novel framework that enables the generation of 3D object prototypes directly from MLLMs, including geometry and part labels. Our pipeline is agentic, comprising a designer, coder, and visual inspector operating in a refinement loop. Notably, our approach requires no additional training data or detailed user instructions. Building on prior work in 2D generation, we demonstrate that rendered images produced by our framework can be effectively used for image classification pretraining tasks and outperforms previous methods by 15\%.
As a compelling real-world use case, we show that the generated prototypes can be leveraged to improve fine-grained vision-language models by using the rendered, part-labeled prototypes to fine-tune CLIP for part segmentation and achieving a 55\% accuracy improvement
without relying on any additional human-labeled data. 
\end{abstract}

\section{Introduction}
\label{sec:introv2}
Large Language Models (LLMs) have proven to be powerful engines for understanding and reasoning in natural language tasks \cite{brown2020language, wang2024mmlu}, and can be used to extract world-knowledge and object properties  \cite{chen2020constructing, abdou2021can, vclm}. While language-only LLMs have been shown to struggle with spatial reasoning \cite{li2024advancingspatialreasoninglarge, cohn2024evaluating, hochmair2024correctness}, recent multimodal LLMs (MLLMs) have had success in spatial discriminative tasks like image understanding \cite{goyal2017making} and 2D visual grounding \cite{kamath2023s}. %

Ultimately, it is desirable that spatial reasoning happens in 3D, as 2D perspectives vary drastically and commonly lead to inconsistent object representations known as the Janus problem~\cite{poole2022dreamfusion}. 
Therefore, we address the following two questions: 1) Is it possible to obtain 3D object prototypes from MLLMs without any additional training data? and 2) Are these prototypes useful for solving downstream vision tasks?
\pagebreak

\begin{figure}[t]
    \centering
    \includegraphics[width=\linewidth]{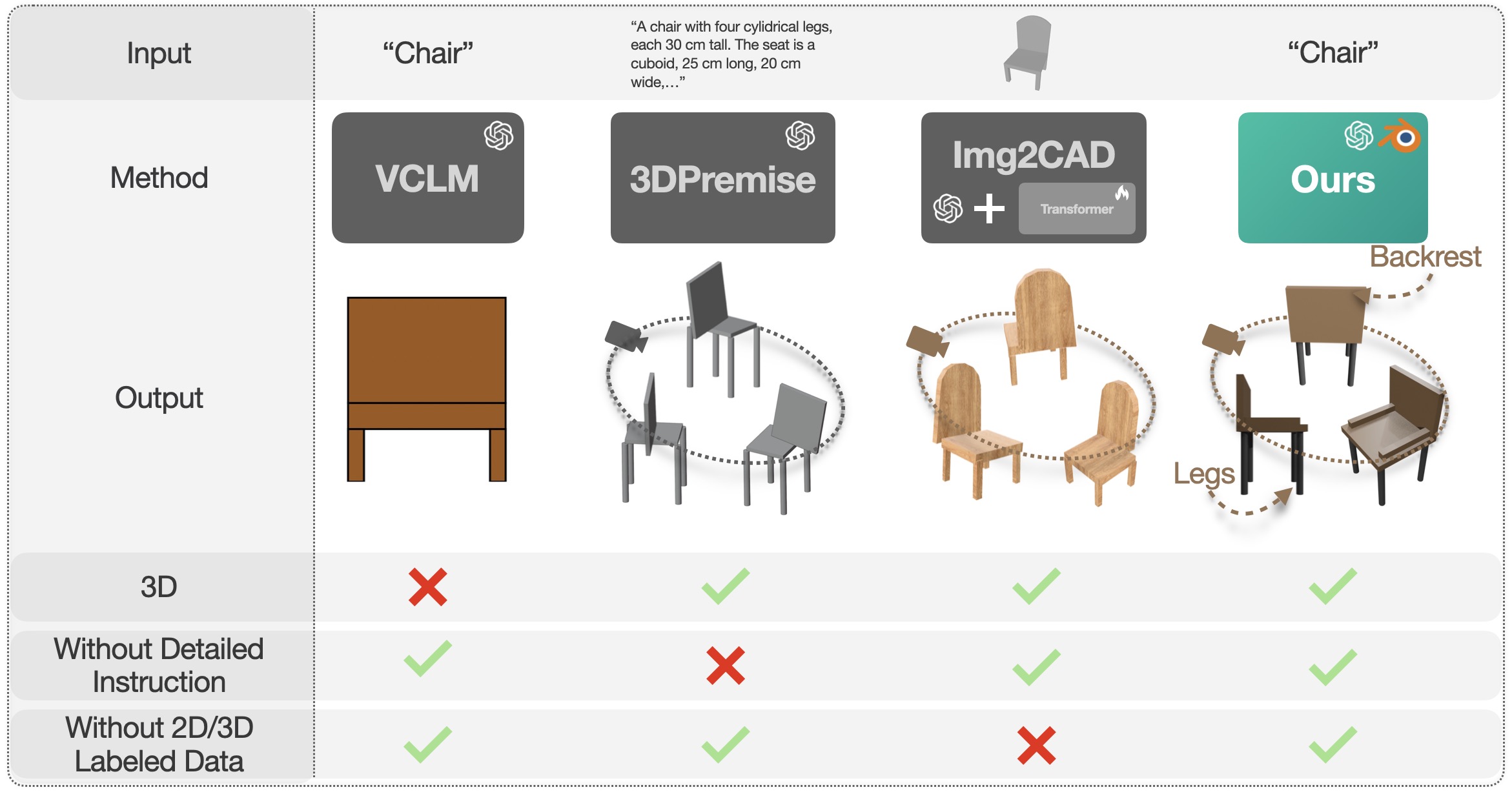}
    \caption[Caption for Teaser]{Previous work can leverage LLMs only to generate 2D view-dependent sketches of objects (left), requires explicit conditioning on detailed human-annotated captions (middle left), or requires additional training data and images as conditioning during inference (middle right\footnotemark). For the first time, we show that 3D object prototypes with their geometry and part labels can be obtained from MLLMs only (right).}
    \label{fig:teaser_fig}
\end{figure}

\footnotetext{Chair image taken from “Img2Cad” \cite{you2024img2cadreverseengineering3d}, used under CC BY-SA 4.0.}

Prior work using LLMs and MLLMs is limited either to 2D, supplying the MLLM with extensive conditioning,  or requiring dedicated task-specific models derived from additional labeled data.
In contrast, for the first time, we show that MLLMs can create 3D object assets with their geometry and part structure on their own (see ~\cref{fig:teaser_fig}). 
These assets, which we refer to as \emph{prototypes}, can be rendered to 2D images with local part annotations and provide a rich signal that can be used to ameliorate the lack of object compositional knowledge in vision-language models (VLMs)~\cite{wei2023ov}.

The foundation of our method is rooted in democratizing 
~\cite{liu2024democratizing} 
actions between three agents: the \emph{\designer{}}, \emph{\coder{}} and the \emph{\visualInspector{}}.
A single agent is responsible for a certain set of sub-tasks to manage complexity, 
and is composed of a sequence of modules. 
This structure effectively manages the complexity of individual tasks, built upon the broad concepts of iterative collaboration and self-refinement \cite{wang2024mixture, liu2024democratizing, zhang2024llm, guo2024large}. 
We make repeated use of the \coder{} and \visualInspector{} by introducing a reconstruction loop where the generated code is rendered, automatically inspected for correctness, and iteratively refined.

To evaluate the quality of our generated prototypes, we first follow the evaluation procedure of the 2D baseline VCLM~\cite{vclm} and generate images that we use as pretraining for image classification. We achieve a $15\%$ %
higher five nearest neighbour classification accuracy, while requiring $35\%$ fewer tokens per generated image. 
Finally, we leverage the 3D part annotations obtained from our method to  show that we can improve CLIP's \cite{radford2021learningtransferablevisualmodels} ability for part-segmentation by $55\%$ over the seminal training-free adaptation method SCLIP \cite{rao2022densecliplanguageguideddenseprediction}. We further ablate our framework and demonstrate improvements in plausibility of part arrangement and overall quality in point cloud registration metrics. In summary, we provide the following contributions:
\begin{itemize}
    \item We present a novel agentic framework for 3D prototype generation from MLLMs, comprising a \designer{}, \coder{} and \visualInspector{} with a feedback loop.
    \item For the first time, we show that generating  3D prototypes from MLLMs including geometry and part structure  is possible without any additional models and training datasets.
    \item We demonstrate the quality of our prototypes by 
    employing rendered images into an image pretraining task 
    and outperform the previous method in this field by $15\%$. 
    \item We finally show that leveraging the part-labels from our prototypes can improve the part segmentation ability of CLIP by $55\%$, while not using any human-annotated data.     
\end{itemize}

\section{Related Work}
\label{sec:litreviewv2}
\myparagraph{2D and 3D Generation using MLLMs.}  
Some previous work explores leveraging LLMs for programmatic generation of diagrams~\cite{belouadi2024automatikz, belouadi2024detikzify} or 2D images~\cite{vclm}. In contrast to the above, our method uses an agentic framework and creates \emph{3D} prototypes.  

Other work explores the generation of 3D scenes and objects from verbose language inputs \cite{sun20243dgptprocedural3dmodeling, delatorre2024llmrrealtimepromptinginteractive, yang2024holodecklanguageguidedgeneration, zhou2024gala3dtextto3dcomplexscene, gao2024graphdreamercompositional3dscene, hu2024scenecraftllmagentsynthesizing, yuan20243dpremiselargelanguagemodels, alrashedy2024generating}. In contrast to the above, our work generates 3D object prototypes solely from \emph{category-name} prompts. 

Finally, some work explores fine-tuning GPT architectures for object synthesis~\cite{siddiqui2023meshgpt, nash2020polygenautoregressivegenerativemodel, yin2023shapegpt3dshapegeneration, du2024blenderllm, you2024img2cadreverseengineering3d, chen2024img2cadconditioned3dcad, wang2024llama}. In contrast, our method does \emph{not require any labeled data or training}.

For baseline comparison, we choose the closely related 2D method VCLM \cite{vclm}. %
Img2Cad \cite{you2024img2cadreverseengineering3d} similarly produces 3D objects with part labels, but works in a different setting as it leverages an additional transformer and training data consisting of 3D models. \\

\myparagraphnoindent{Improving Spatial Reasoning in MLLMs.}
Prior works \cite{weston2015towards,mirzaee2021spartqa,shi2022stepgame} have assessed MLLM's spatial reasoning abilities, and have found that they struggle with complex spatial reasoning tasks when they are presented naively in a single question/answer round. Recent methods \cite{li2024advancingspatialreasoninglarge} address these shortcomings by incorporating prompting strategies like Chain-of-Thought (CoT) reasoning \cite{wei2022chain}. Therefore, we incorporate CoT in our \designer{} and \coder{} agents \cite{durante2024agentaisurveyinghorizons} to improve performance in the 3D generation domain. 
Another useful technique for LLM querying is \emph{self-refinement} \cite{madaan2023selfrefineiterativerefinementselffeedback}, where an LLM is instructed to improve its own responses, analogous to how humans refine ideas through iterative exploration. It has been successful in domains including object recognition \cite{wu2024self}, image design \cite{yang2023idea2img}, and code generation \cite{dong2024self}. Furthermore, \cite{gou2023critic} argue that self-refinement solely with LLMs can lead to hallucinations, and introduce CRITIC, which allows LLMs to interact with external tools for validation and correction. Inspired by these works, our agents are self-refining, and have access to an external graphics engine (Blender \footnote{https://www.blender.org/})  
to render views that are then fed back into the MLLM. 

\myparagraphnoindent{Generating Visual Data from Language.} 
Increasing emphasis is being put on leveraging LLMs in creating synthetic training data for downstream vision tasks, with the goal of scaling data creation, reducing the need for costly manual annotation and improving performance. VCLM \cite{vclm} found that language models learn visual properties during training which can be translated to 2D sketches in order to supervise visual models. PIXMO \cite{deitke2025molmo} generates renders by writing graphical programs in different programming languages,
which are then used for training a VLM to improve its visual reasoning skills, though the application is limited to chart and diagram understanding. DALDA \cite{jung2024dalda} proposes an LLM-in-the-loop framework that synthesizes training images with a diffusion model. However, it suffers from view-dependent generation. 
In contrast, our approach generates synthetic data for downstream vision tasks without additional training, and produces 3D object geometries including part annotations. This enables more diverse and structured data generation, supporting  fine-tuning of CLIP for part segmentation, which is an important real-world application that CLIP inherently struggles with \cite{wei2023ov}.

\begin{figure*}[t]
  \centering
  \includegraphics[width=\textwidth]{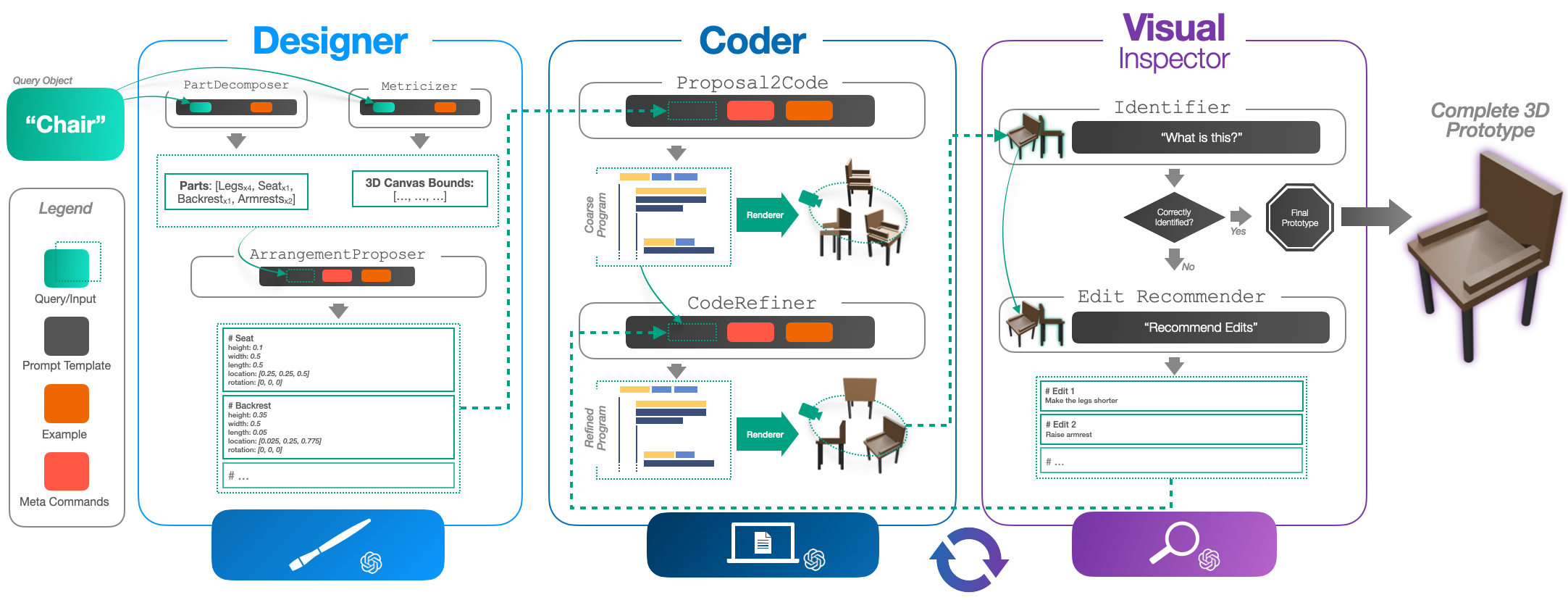}  
  \caption{
  \textbf{
  Overview of \method{}}. 
  We propose a novel 3D prototype generation framework from MLLMs without the need of detailed user instructions or additional training data.
  The figure visualizes the generation  for the example query ``Chair" (left) to a 3D model (right). The framework delegates actions between three agents to manage task complexity: the {Designer} is responsible for generating a 3D canvas and extracting object compositional knowledge. The {Coder} converts it into renderable code and finally the {Visual Inspector} works in a reconstruction loop with the {Coder} to iteratively improve the model. 
  \vspace*{-3mm}
  }
  \label{fig:main_fig}
\end{figure*}

\section{Methodology}

\subsection{Pipeline Overview}

While instances in the real-world may have varied dimensions, part arrangements, and appearances, this work studies generating a typical instance that we refer to as a \emph{3D prototype}, which captures the  overall 3D geometry and part layout.  
For instance, a possible prototype for a  chair would consist of four cylindrical legs supporting a flat seat with a flat backrest, arranged with typical proportions and spatial relationships. 
We aim to generate these prototypes only from MLLMs and establish a framework consisting of three agents~\cite{durante2024agentaisurveyinghorizons}: a \designer{} 
for object decomposition, a \coder{}  
to take the decomposition to produce renderable 3D programs, and a \visualInspector{} 
that can provide feedback to iteratively refine the 3D model in conjunction with the \coder{}. Each agent is subdivided into modules that are responsible for solving certain subtasks though prompting the MLLM. An overview is provided in~\cref{fig:main_fig}. In the following, we present the high-level building blocks of our framework and refer to the supplementary material for implementation details.

\subsection{Designer}
\label{subsec:designer}
We follow~\citet{ranasinghe2024learning} who suggest that a two-stage approach of first planning and then localizing results in more accurate spatial reasoning with MLLMs. To this end, we structure our Designer agent into 3D \emph{canvas generation} (planning) and \emph{arrangement proposal} for individual parts (localizing).
\\

\myparagraphnoindent{Canvas Generation.}
Similar to how human artists internalize the visual bounds of their sketches, our method first extracts the 3D bounds in which the prototype will be generated. More specifically, given the desired category name as the query input $q$, a \texttt{Metricizer} extracts the 3D canvas bounds:

\[ C \leftarrow{} \texttt{Metricizer}(q) \,, \]  
with $C \in \mathbb{R}^3$. By ensuring that the prompt encourages the MLLM-Engine to produce a canvas in metric as opposed to pixel units, we posit that the \texttt{Metricizer} is able to tap directly into the world-knowledge of the MLLM during traning about object dimensions obtained from manuals, CAD programs and product specification documents. 

Concurrently, the task of object decomposition is delegated to to the \texttt{PartDecomposer} that produces a set $D$ of part-labels:

\[
    D = \{(l_1, c_1),....,(l_n,c_n)\} \leftarrow{} \texttt{PartDecomposer}(q) \,, 
\]
where $l_i$ refers to the part label and $c_i$ 
refers to the expected instance count for the corresponding part. We find that for common objects this naive decomposition works very reliably and avoids hallucinations \cite{huang2025survey}.

\myparagraphnoindent{Arrangement.}
Next, we introduce the \texttt{ArrangementProposer} that uses $C$ and $D$ to reason about the arrangement of individual parts: 
\begin{align*}
A &= (R_1, \dots, R_n) \\[5pt]
    &\leftarrow \texttt{ArrangementProposer}(q, D, C) \,, 
\end{align*}
where $R_i = \{\mathbf{d}_i, \mathbf{p}_i, \boldsymbol{\theta}_i\}$ for each part $i$ refers to a parametrized cuboid defined by 3D dimensions $\mathbf{d}_i$, 3D position $\mathbf{p}_i$, and a 3D rotation $\boldsymbol{\theta}_i$. 

\subsection{Coder}
\label{subsec:coder}
The \texttt{Coder} begins by explicitly representing $A$ into a program in a language and modeling API $L$,  
where the choice of $L$ dictates the rendering tool that will be used later (e.g. Blender). 
We define the first module in this agent as \texttt{Proposal2Code}:
\[
    {P}_{Coarse} \leftarrow{} \texttt{Proposal2Code}(A, L) \,,
\]
where $P_{Coarse}$ represents the first coarse program. Next, the \texttt{CodeRefiner} reasons about ${P}_{Coarse}$ and directly generates a refined program:
\[
    {P}_{0} \leftarrow{} \texttt{CodeRefiner}( {P}_{Coarse}, L) \,. 
\]
The intuition is that by first creating a coarse program, we can reduce the problem space for the \texttt{CoderRefiner} to focus on the fine-grained modeling details of the 3D model. As most languages provide functions for primitives (i.e. cylinder, cube, torus, etc.), 
we encourage the model towards using these primitive functions through our in-context examples.
Given that GPT-4o was able to reliably produce error-free programs, we avoid complex methodologies like RAG querying \cite{lewis2020retrieval}. 

${P}_{0}$ is now rendered using a rendering engine (\texttt{Renderer}) to obtain three views randomly sampled from the upper-hemisphere of the object:

\[
    {R}_{\texttt{0}} \leftarrow{} \texttt{Renderer}({P}_\texttt{0}) \,
\]

\subsection{Visual Inspector}
\label{subsec:visual_inspector}
With the {Visual Inspector}, we intend to criticize the 3D prototypes produced by the {Coder} and iteratively improve them.
According to \citet{liu2024mibench}, the recent GPT-4o shows emergent properties in reasoning over multiple images as input. To increase robustness, we therefore input all three renderings into our {Identifier}, which asks the MLLM to predict the object from the images:
\[
    \hat{q}_0 \leftarrow{} \texttt{Identifier}(R_{\texttt{0}}) \,, 
\]
where we stop any further refinement if the predicted $\hat{q}_0$ is equal to the original query $q$ and take ${P}_{{0}}$ as the final output. 

\subsection{Self-Refinement/Reconstruction Loop} 

As long as the \texttt{Identifier} is unable to predict the input query, the processing continues with the module \texttt{EditRecommender}. Given the previous renders, the incorrect predictions, and the input query, the module queries the MLLM engine to recommend natural language edits:
\begin{align*}
E_0& = (e_{1}, \dots, e_{m}) \\[5pt]
    & \leftarrow \texttt{EditReccomender}({R}_{\texttt{0}}, q, \hat{q}_0) \,, 
\end{align*}
where $e_i = (v_i, \tau_i, w_i)$ refers to a set of proposed edits defined by the visual aspect $v_i$ represented as a string (e.g.\ ``tire of the car”), the general category of edit $\tau_i$ (like {delete, rotate, move}, etc.), and the  editing command in natural language  (e.g.\ ``move the tires of the car to the corners of the base"). The recommended edit is then used to refine the prototype with the \texttt{CodeRefiner}:
\[
    {P}_{k} \leftarrow{} \texttt{CodeRefiner} ({P}_{k-1}, E_{k-1}, L) \,.
\]
Given a refinement budget of three iterations, we iterate between the \texttt{CodeRefiner} and the Visual Inspector until the iterations are exceeded 
or the stopping criteria implemented through the \texttt{Identifier} is met. 
To match our baseline's setup, the final result is accepted even when the iteration budget is exceeded.

\section{Experiments}

\subsection{Visual Pretraining}
\label{subsec:vis_pre}

\subsubsection{Datasets and Baselines.}
To evaluate the quality of our generated prototypes, we follow the protocol of VCLM~\cite{vclm} and leverage renders of our prototypes to pretrain object classification models. As their work generates 2D sketches from synthesized programs, we establish the following baseline variants: 

\textbf{$\text{VCLM}_{\text{GPT-3.5}}$} refers to a set of 31k 2D sketches created using GPT-3.5. \textbf{$\text{VCLM}_{\text{GPT-3.5+GPT-4}}$} is a larger dataset with 80k images that combine the original set with additional sketches generated by GPT-4. Finally, we run their prompting scheme with the MLLM GPT-4o and refer to it as  \textbf{$\text{VCLM}_{\text{GPT-4o}}$}.

To benchmark our prototypes against these baselines, we first create a dataset of 3D prototypes with our method using GPT-4o as well. We use the top 300 most frequent classes in ADE20K~\cite{zhou2019semantic} as input queries. Similar to our baseline, we condition our \coder{} agent to synthesize code from three different languages and modeling APIs:
the Python API for Blender, Processing 3D in Java, and Matplotlib 3D.
The images rendered during the refinement process yield the final set of 80k images labeled as \textbf{$\text{3DFroMLLM}_{\text{GPT-4o}}$}. 

\subsubsection{Qualitative Evaluation.}
\begin{figure*}[t]
    \includegraphics[width=\textwidth]{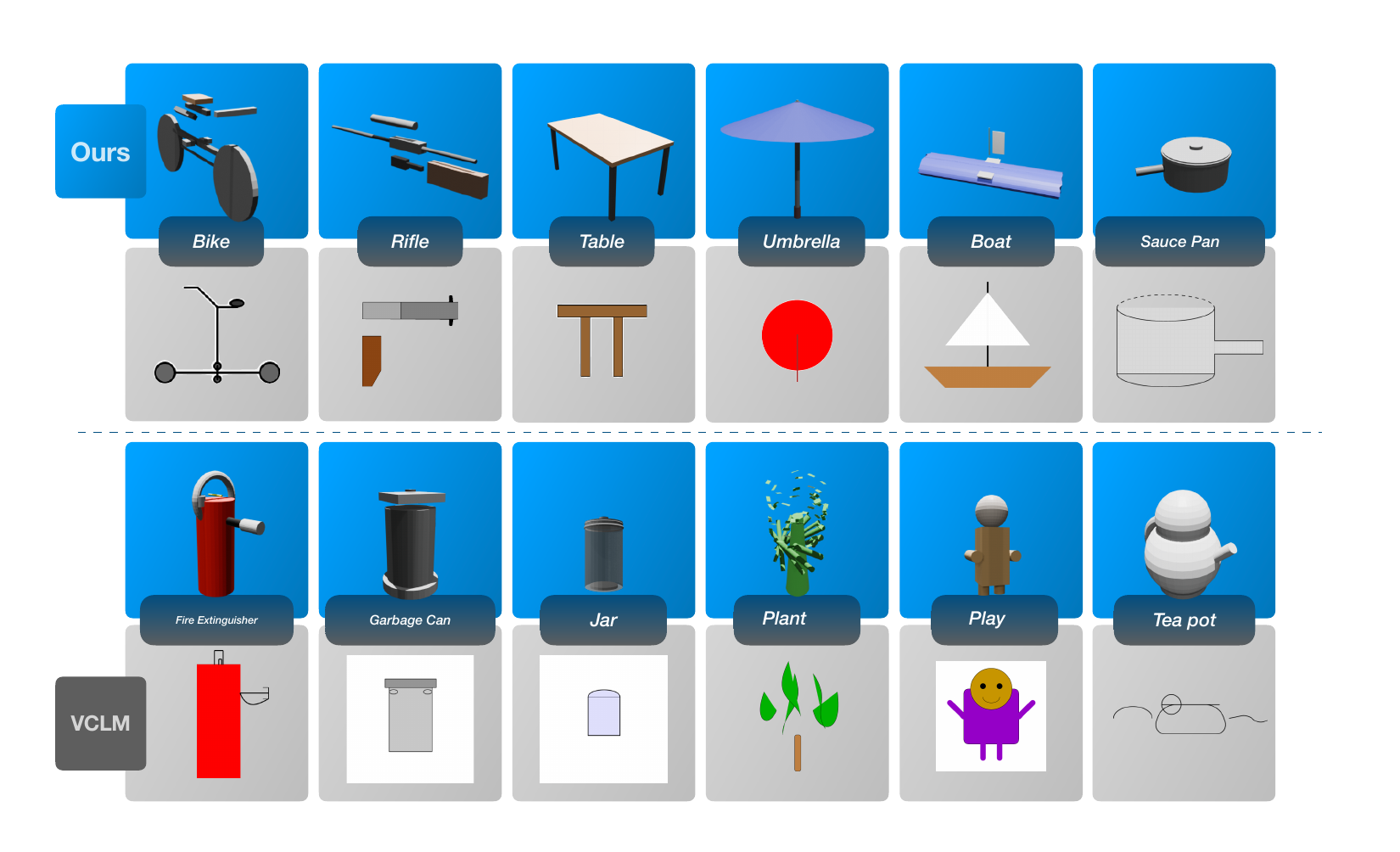}
    \caption{\textbf{Qualitative comparison of our method and the baseline $\text{VCLM}$} both executed with with ${\text{GPT-4o}}$. While a single instance from the baseline (gray background) produces a single 2D view, one of our prototypes can be rendered from arbitrarily many views. 
    As visible, our 3D prototypes are of a significantly higher level of detail and the renderer provides shading cues that give a clearer impression of the geometry. 
    \vspace*{-3mm}
    }
    \label{fig:qualitative}  
\end{figure*}

In \cref{fig:qualitative}, we provide a qualitative comparison of our generated prototypes to the VCLM baseline. As visible, our method is able to generate 3D models for the prompts successfully with our prototypes showing a significantly higher level of detail. In addition, the renderer is able to generate shading cues that lead to a significantly higher degree of realism. A preliminary investigation on how shading cues effect downstream pretraining performance is provided in the supplementary. In~\cref{fig:variation}, we furthermore compare the ability to generate different variants and how our framework can simply be extended to handle requested edits.

\begin{table}[h]
\centering
\begin{tabular}{llcl}
    \specialrule{.2em}{.1em}{.1em}
    & \multirow{1}{*}{Pre-training} & \multicolumn{1}{c}{Tokens per} &\multicolumn{1}{c}{I-100} \\
    & Dataset & image & 5-NN     \\
    \specialrule{.1em}{.05em}{.05em}
    Random & None & - & 4.28    \\
    \cmidrule{2-4}
    \multirow{3}{*}{2D Baselines} & $\text{VCLM}_{\text{GPT-3.5}}$  &     &   22.42        \\ 
    & $\text{VCLM}_{\text{GPT-4+3.5}}$    &  & 27.44          \\ 
    & $\text{VCLM}_{\text{GPT-4o}}$ & 294 & 29.68\\
    \cmidrule{2-4}
    \multirow{1}{*}{Ours} & $\text{3DFroMLLM}_{\text{GPT-4o}}$ & \textbf{191} & \textbf{34.02} \\

\specialrule{.1em}{.05em}{.05em}
\end{tabular}
\caption{\textbf{Pretraining on generated datasets}. The table shows the 5-nearest-neighbor classification accuracy  on ImageNet-100 of the pretrained models. All models significantly distinguish themselves from a randomly initialized one (row 1). Pretraining VCLM with the more powerful GPT-4o yields improved performance (row 4), 
 while pretraining on renders from our rendered 3D prototypes outperforms all baselines (last row). Notably, our method requires only 65\% of the compute (tokens). 
}
\label{tab:moco_performance}
\end{table}
\begin{table}[!]
\centering
\begin{tabular}{lll}
\specialrule{.2em}{.1em}{.1em}
& \multirow{2}{*}{Pre-training Dataset} &\multicolumn{1}{c}{I-100} \\
& & 5-NN     \\
\specialrule{.1em}{.05em}{.05em}
Real & Places &   57.04    \\
\cmidrule{2-3}
\multirow{4}{*}{Procedural} & Dead-leaves         & 12.76        \\ 
& FractalDB-1k        & 17.24 \\ 
& StyleGAN O          &     33.00    \\
& S-21k               &  43.24  \\ 
\cmidrule{2-3}
\multirow{3}{*}{2D Baselines} & $\text{VCLM}_{\text{GPT-3.5}}$ + S-21k  & 43.40  \\ 
& $\text{VCLM}_{\text{GPT-4+3.5}}$  + S-21k  & 43.36  \\ 
& $\text{VCLM}_{\text{GPT-4o}}$ + S-21k & 55.86 \\ 
\cmidrule{2-3}
Ours & $\text{3DFroMLLM}_{\text{GPT-4o}}$ + S-21k & \textbf{56.20} \\

\specialrule{.1em}{.05em}{.05em}
\end{tabular}
\caption{\textbf{Pretraining on mixup with textured synthetic datasets.} 
The table shows the 5-nearest-neighbor classification accuracy  on ImageNet-100 of the pretrained models. 
Images from our newly trained baseline $\text{VCLM}_{\text{GPT-4o}}$ (row 8) combined with the  procedurally generated texture dataset Shaders21k (S-21k) significantly outperform the previous versions (rows 6 and 7).
Our method performs best (row 9) and achieves a performance that is remarkably close to when pretraining models on real data (row 1). 
} 
\label{tab:moco_performance_textured}
\end{table}

\begin{figure}[t]
    \includegraphics[width=\linewidth]{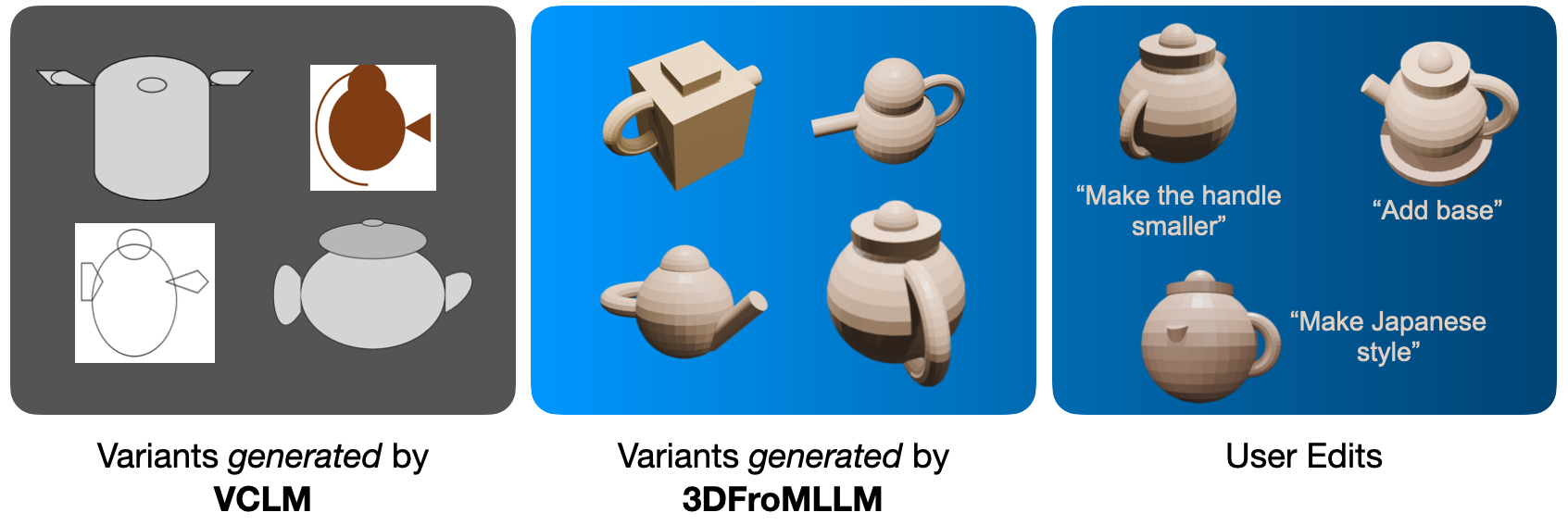}
    \caption{
    \textbf{
    Left: variants generated by VCLM. Middle: variants generated by our method.} As visible, our method is able to generate significantly more plausible variants. \textbf{Right: edits performed on our models through user commands.}
    }
    \label{fig:variation}
\vspace{-3mm}
\end{figure}

\myparagraphnoindent{Evaluation Protocol.}
To measure the usefulness of our image dataset  $\text{3DFroMLLM}_{\text{GPT-4o}}$ and the baseline $\text{VCLM}_{\text{GPT-4o}}$ in learning vision backbones, we follow the well-established training protocol from \citet{baradad2022procedural} to train a ResNet-50. 
In the beginning, all datasets are expanded using a Dirichlet mixup strategy~\cite{zhang2017mixup, baradad2021learning},  to a size of 1.3M images with resolution $384 \times 384$. 
Next, for each dataset, we pretrain the ResNet using the MoCo-v2 method for 200 epochs with batch size of 256. After the pretraining, we evaluate the performance on the ImageNet-100 dataset by evaluating the 5-nearest-neighbor classification accuracy following~\citet{vclm}. We report the classification accuracy performance in~\cref{tab:moco_performance} and notably outperform the baseline by a large margin in terms of both, classification accuracy and required compute.

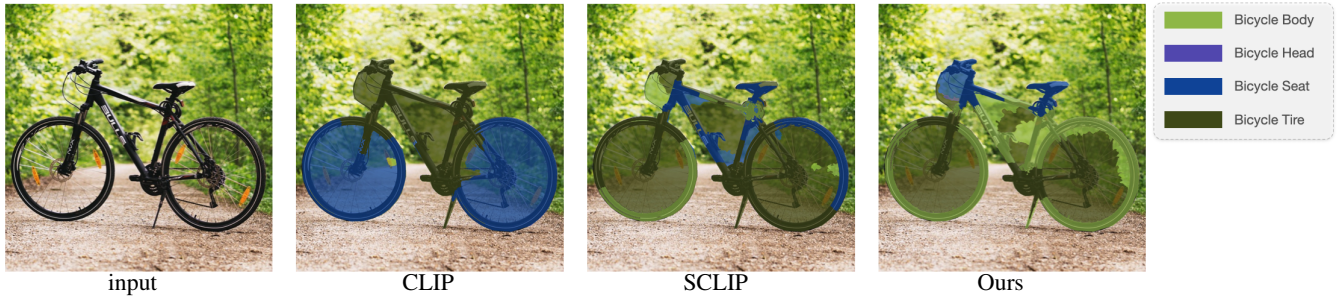
\begin{figure*}[t]
    \begin{overpic}[width=\linewidth]{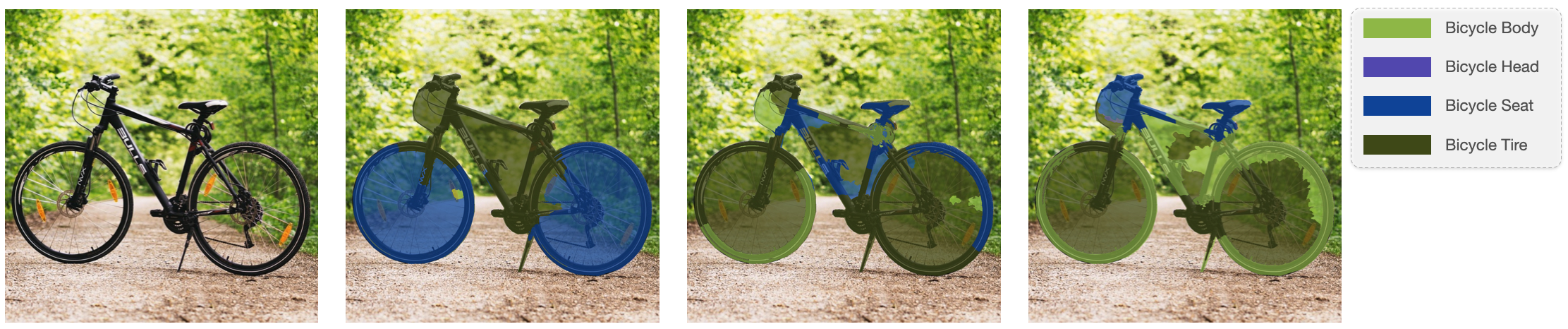}
        \put(8,-1){\small input}
        \put(30,-1){\small CLIP}
        \put(51,-1){\small SCLIP}
        \put(73,-1){\small Ours}
        
    \end{overpic}
    
    \caption{Proposed segmentation map of bicycle parts from CLIP and SCLIP before and after finetuning on our part-labeled prototypes. Our finetuned CLIP (rightmost) can better localize parts with significantly lower noise. Labels in the background were removed for visualization purposes. 
    \vspace*{-4mm}
    }
    \label{fig:segmentations}
\end{figure*}

\myparagraphnoindent{Experiments with Textured Mixup.}
We first evaluate the pretraining on a set of different procedurally generated textures for which we leverage the methods Dead-Leaves and Style-GANo from \citet{Kataoka_2020_ACCV}, FractalDB-1k from \citet{baradad2021learning}, and Shaders21k \cite{baradad2022procedural}. We find that  Shaders21k yields the highest performance (see top of~\cref{tab:moco_performance_textured}). Subsequently, we explore using mixup of of Shaders21k textures and the generated protoypes from VCLM and our method (see bottom of~\cref{tab:moco_performance_textured}). We generally note that using the multi-modal GPT-4o model for the data generation increases the performance significantly and brings it close to when pretraining on real data. 
We outperform the VCLM baseline and achieve performance close to pretraining on real datasets.

\subsection{Improving Part Segmentation in VLMs}
\label{subsec:clip_experiment}

\begin{table}
  \centering
  \begin{tabular}{llS}
  \specialrule{.2em}{.1em}{.1em}
  & \multirow{2}{*}{Fine-tuning Dataset} & mIoU \\
  & & $\text{PartImageNet}_{\text{test}}$ \\
  \specialrule{.1em}{.05em}{.05em}
  \rowcolor{gray!20}
  CLIP & $\text{PartImageNet}_{\text{train}}$ & 51.95 \\
  \cmidrule{2-3}
  CLIP & - & 4.90 \\ 
  SCLIP & - & 10.89 \\ 
  \cmidrule{2-3}
  CLIP & 3DFroMLLM+CN & \textbf{16.90} \\
  SCLIP & 3DFroMLLM+CN & 10.92 \\ 
  \specialrule{.1em}{.05em}{.05em}
  \end{tabular}
  \caption{\textbf{Finetuning CLIP for part-segmentation.} 
   The Top shows the upper bound when a human-labeled dataset is used. The second and third rows show CLIP and SCLIP performance  for part segmentation without any finetuning and the fourth and fifth row after finetuning on our data. Notably, our data helps to improve part segmentation for CLIP by a large margin, without using additional human-labeled datasets. 
  } 
  \label{tab:clip_eval}
\end{table}

\myparagraph{Motivation.} 
\citet{wei2023ov} found that CLIP has difficulty segmenting object parts, which is a critical ability for any downstream application that involves content understanding or robotics. 
Our method offers a new source of object part annotations without requiring additional human labeling. To this end, we extend the real-world utility of our prototypes beyond visual pretraining (\cref{subsec:vis_pre}) by fine-tuning CLIP to improve performance on part segmentation. 
\\

\myparagraphnoindent{Dataset.} Our prototypes can be used as guidance to produce naturalistic images of objects with ground-truth part segmentations, by leveraging them to condition diffusion-based image generation. We first generate a single prototype for each of the 11 super-categories in PartImageNet \cite{he2022partimagenetlargehighqualitydataset} and then render depth maps and segmentation masks. We subsequently leverage ControlNet \cite{zhang2023addingconditionalcontroltexttoimage} to condition StableDiffusion~\cite{rombach2022high} on the depth maps, 
in order to create naturalistic images.
This yields a dataset of 20.4k images and part segmentation masks that we refer to as $\text{3DFroMLLM+CN}$. Evaluation is performed on the test split of PartImageNet, which contains 2408 images with 67 unique part labels. Additional details are provided the supplementary material. \\

\myparagraphnoindent{Experimental Protocol.} We finetune only the visual backbone of the CLIP model by supervising the pixel-text score maps. Only the last transformer layer is trained. 
We compare our finetuned CLIP against the standard CLIP and SCLIP.
As an upper-bound, we report the performance of CLIP when finetuned using real ground-truth image-mask pairs ($\text{PartImageNet}_{\text{train}}$). \\

\myparagraphnoindent{Results.} In~\cref{tab:clip_eval}, we present mIoU scores on the PartImageNet testing split. We note that CLIP trained using our MLLM-generated prototype dataset outperforms both the original CLIP and the specialized SCLIP, exceeding it by about 55\%. We provide a visual example in \cref{fig:segmentations}. Although the current results remain far from the upper-bound, our method demonstrates more effectiveness than conventional methods that follow training-free adaptation.

\begin{table*}[t]
  \centering
  \begin{tabular}{c c c c c c c c}
      \specialrule{.2em}{.1em}{.1em}
      \multirow{2}{*}{\textbf{Method}} & 
      \multicolumn{3}{c}{\textbf{Agents}} & 
      \textbf{Sparse} $\downarrow$ &
      \multicolumn{2}{c}{\textbf{Dense} $\downarrow$} \\
       & Designer & Coder & Visual Insp.                & 5NN    & Chamfer & Hausdorff \\
      \specialrule{.1em}{.05em}{.05em}
        & × & × & × & \underline{0.044} & 0.233 & 0.515 \\
      \cmidrule{2-7}
      \multirow{4}{*}{\method{}} & \checkmark & × & ×   & 0.057 & 0.255 & 0.631 \\
      \cmidrule{2-7}
       & \checkmark & \checkmark & ×                    & 0.056 & 0.223 & 0.572 \\
       \cmidrule{2-7}
       & \checkmark & × & \checkmark                    & 0.055 & 0.200 & 0.520 \\
       \cmidrule{2-7}
       & × & \checkmark & \checkmark                    & 0.062 & \textbf{0.159} & \textbf{0.448} \\
      \cmidrule{2-7}
       & \checkmark & \checkmark & \checkmark           & \textbf{0.041} & \underline{0.181} & \underline{0.453} \\
      \cmidrule{1-7}
      \rowcolor{gray!20}
      $\text{Real}^{\text{3D}}$ & - & - & - & 0.052 & 0.065 & 0.163 \\
      \specialrule{.1em}{.05em}{.05em}
  \end{tabular}
  
  \caption{
  \textbf{Ablation study over agents.} 
   The sparse evaluation metric focuses on part arrangement plausibility and compares only the part centroids of generated prototypes to a reference dataset over the five nearest neighbors, while the dense metric evaluates the best matching geometries.  As visible, using either the Coder and Visual Inspector, or all of our agents yields best results. The final choice depends on the application. As for for the refinement of VLMs we focus on part arrangement, we prioritize the sparse metric and use all agents. 
  \vspace*{-3mm}
  }
\label{tab:part_arrangement}
\end{table*}

\begin{figure*}[t]
  \centering
  \includegraphics[width=0.7\textwidth]{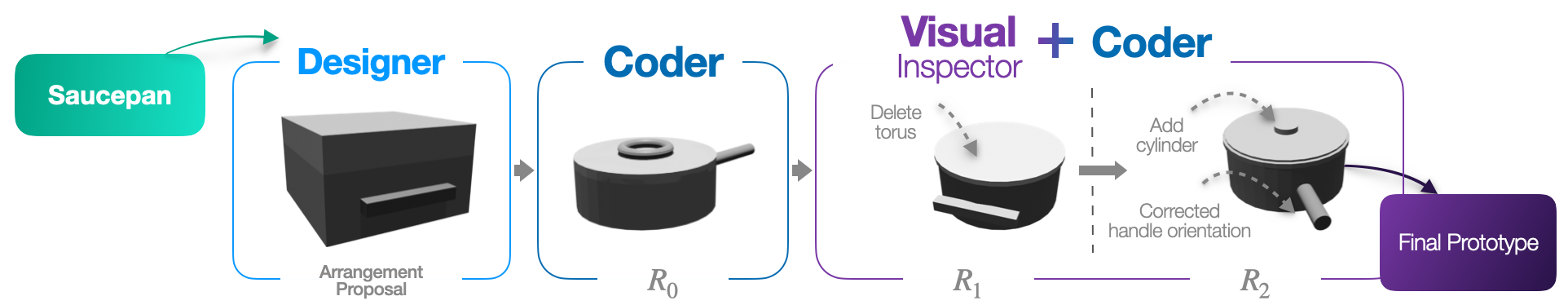}
  \caption{\textbf{A qualitative example showing the effect of our agents} when queried to generate a `Saucepan'.
  The Designer proposes an initial box layout, while the coder refines it with details in $R_1$. Note that the Torus is not realistic. While quality temporarily decreases in the first refinement iteration, the Visual Inspector is eventually able to produce a model with higher level of detail and accuracy (top handle and lid). 
  \vspace*{-4mm}
  }
  \label{fig:self_correction}
\end{figure*}

\subsection{Ablation} 
\label{subsec:ablation}
\myparagraph{Dataset.}
To measure the contribution from each agent in our framework, we propose to compare the sparse and dense geometry of our prototypes
against the reference human-authored CAD dataset PartNet~\cite{Mo_2019_CVPR}. 
To make the evaluation computationally feasible, we sample 500 instances per category for each of the 47 categories, which we refer to as \textit{targets}. Subsequently, we generate a prototype for each of the 47 categories with our method and compare it to the targets for the category respectively. 

\myparagraphnoindent{Protocol for Sparse Part Arrangement.}
We propose to evaluate the plausibility of the part arrangements by summarizing each part to its centroid and comparing the centroid positions to the reference dataset. 
To find the correspondences between part centroids from our prototypes and the targets, for a given prototype-target pair, we request GPT-4o to generate a mapping between the prototype part labels to the closest target part label. Parts that are left unmapped are disregarded in further computation.
We then align the prototype to the target by determining an optimal transformation using the Kabsch-Umeyama algorithm \cite{Kabsch}. 
With this, we compute all prototype-target part distances. To avoid outliers due to the sub-sampling, and to increase robustness of the evaluation, the final sparse metric is computed as the average of the five nearest targets.
The pseudo-code is provided in the supplementary material. 

\myparagraphnoindent{Protocol for Dense Geometric Evaluation.}
In contrast to the above, the dense evaluation measures how closely the prototype geometry follows the human-authored CAD models exactly.
For each generated prototype, we uniformly sample 10k surface points using the Blender API. This is compared against the point clouds for each target that are already available in the PartNet dataset. 
Following the dense alignment protocol of \citeauthor{alrashedy2024generating}, we transform the prototype to the reference frame of the target with the Iterative Closest Point (ICP) algorithm \cite{glira2015a} and then measure the Chamfer and Hausdorff distance between the aligned prototype and target pair. 

\myparagraphnoindent{Results.} We present results for different combinations of agents in~\cref{tab:part_arrangement}. 
We include a non-agentic baseline (top row), which directly produces Blender code (details provided in the supplemental).
For a holistic view, we also report an upper bound performance when using prototypes from the human-authored $\text{Real}^{\text{3D}}$ dataset instead of our method.

Notably, the agent combination of \designer{}+\coder{} (3rd row) produces worse geometric results than the non-agentic baseline. This is an expected outcome, since the \designer{} produces only a coarse cuboid-like object and the \coder{} initially only focuses on primitive placement, which is also supported by the example in~\cref{fig:self_correction}.  
The \coder{}+\visualInspector{} combination produces the best geometry (Chamfer distance of 0.159), although often at the cost of arrangement plausibility (5NN distance of 0.062). 
Which option to choose depends on the application. As the plausibility of the part arrangement is critical for VLM refinement, we choose the option with all agents.

\section{Conclusion}

Our work is the first to show that it is possible to obtain 3D object prototypes including their geometry and part structures solely from MLLMs,  without any detailed user instructions or additional human-labeled training data. 
We have demonstrated the usefulness of our prototypes by leveraging them to pretrain classification models and outperforming previous methods. Finally, our prototypes can be used to significantly improve the the part-segmentation accuracy of CLIP, enabling more fine-grained vision-language models for important real-world applications that involve content understanding or robotics.

\bibliography{aaai2026}

\end{document}


\maketitle

\section{Rendering Engine and Function}

As described in the main text the choice of rendering function is dictated by the choice of \text{PMA}. 

\myparagraph{Blender} \footnote{www.blender.org} For the blender rendering engine we create a procedural rendering function which programmatically does the following:

\begin{itemize}
    \item Create an icosphere primitive of relative scale around the prototype  
    \item Place Cameras at each vertex
    \item Randomly light the scene
    \item Render the views with perspective projection
\end{itemize}

While the blender rendering engine in the most versatile, we also create rendering engine for the other two \texttt{PMA}s mentioned in . 

\myparagraph{Processing3D} \footnote{www.processing.org} sketches, generated \texttt{sketches} were parsed to first remove unscoped transformations. Followed by adding the following two lines to the \texttt{draw} function and a renderImage function which renders images at each call for draw:

\begin{tcolorbox}[colback=gray!10, colframe=black, sharp corners=southwest, fonttitle=\bfseries]
    \texttt{translate(width/2, height/2, 0);\\
    rotateY(rotation);}
\end{tcolorbox}

This allows a camera orbiting behavior which is utilized to then render several images. 

\myparagraph{Matplotlib 3D} \footnote{www.matplotlib.org} although having a wire-frame look, matplotlib code was reliably produced with renderable shapes and objects. We run procedural code in a thread and capture axes of type \texttt{Axes3D} and simply define a rendering function to produce figures at uniform azimuth increments.

\myparagraph{Failsafes.} Given, that our agents talk to each other, we also implement fall-back mechanisms in the case when programs do not render. Namely, simply resubmitting the previous API request can often ameliorate simpler errors.

\section{Diversity (LPIPs) Score against VCLM}
To supplement the qualitative figures from our method, we additionally report the LPIPs diversity score in \cref{tab:lpips}, quantitatively demonstrating that our dataset exhibits higher perceptual variation than the baseline. \textit{LPIPs} score is measured as the mean pair-wise perceptual variance over the complete datasets using VGG as the feature extractor.

\begin{table}
    \centering
    \begin{tabular}{l | S[table-format=1.3]}
        \textbf{Method} & {LPIPs Score} \\
        \midrule
        $\text{VCLM}_\text{GPT-4o}$ & 0.522 \\
        $\text{3DFroMLLM}_\text{GPT-4o}$ & \textbf{0.551} \\
    \end{tabular}
    \caption{Diversity scores for the 2D renders generated from our method and the baseline, showing better diversity across renders for our method.}
    \label{tab:lpips}
\end{table}

\section{Implementation Details for CLIP Finetuning Experiment}
\subsection{Dataset Generation}
\begin{figure}[!]
    \centering
    \includegraphics[width=\linewidth]{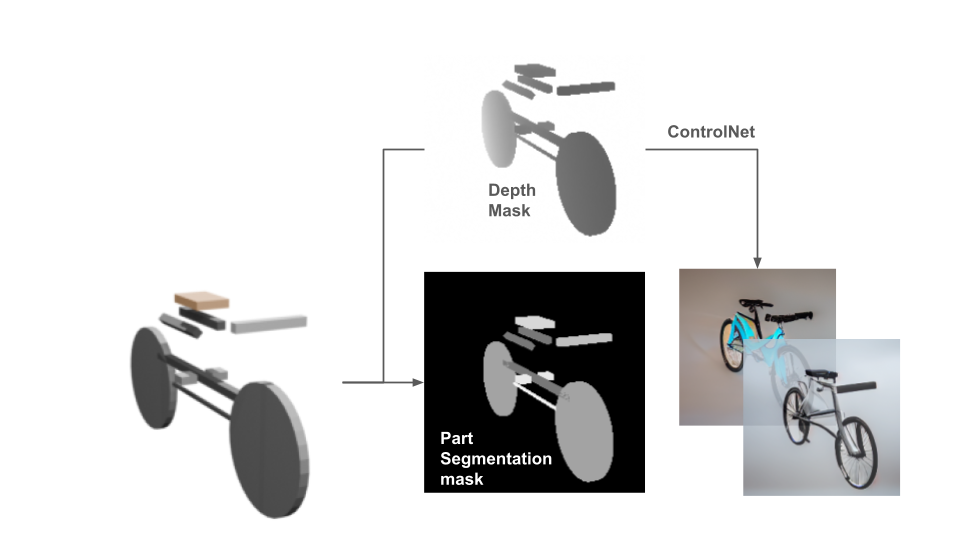}
    \caption{A flow diagram of the CLIP-finetuning data generation process starting from a bicycle prototype.}
    \label{fig:clip_flow}
\end{figure}

For reliable reproduction of our CLIP experiment, in this section, we further provide implementation details. To remain comparable with the training split of PartImageNet we produce a exactly 20,466 mask-image pairs. To achieve this, we create a prototype for each of thee 11 PartImageNet category. We then render each of the 11 prototypes from 88 unique, uniformly placed camera views. This produces a slight excess of mask-image pairs, which are simply randomly filtered out. A depth map rendered from each view is passed to ControlNet (v1.1 \cite{zhang2023addingconditionalcontroltexttoimage}), which provides conditioning to StableDiffusion 1.5 \cite{rombach2022high} to generate 24 realistic samples of size 384x384. The simple prompt ``a photo of a..." was used as the textual prompt while the negative prompt was fixed as ``lowres, bad anatomy, bad hands, cropped, worst quality''. An example of the data generation flow may be found in \cref{fig:clip_flow}. Further hyperparameters for ControlNet can be found below:

\begin{tcolorbox}[colback=gray!10, colframe=black, sharp corners=southwest, title=\textbf{ControlNet Hyper Parameters}, fonttitle=\bfseries]
\texttt{ddim\_steps=20\\
guess\_mode=False\\
strength=1.0\\
scale=9.0\\
seed=12345\\
eta=1.0}
\end{tcolorbox}

\subsection{CLIP Fine-tuning}
For fine-tuning CLIP, we take a pretrained VIT-B-16 CLIP model. We find that fine-tuning only the final layer of the CLIP Vision Transformer leads to the best results. By treating the pixel-text score maps as segmentation results, we can supervise the model directly to produce dense predictions. This training paradigm is simply the established regimen as proposed by DenseCLIP \cite{rao2022densecliplanguageguideddenseprediction}, albeit without a decoder and context-aware prompting. We do this to remain comparable, in terms of number of parameters, to the original CLIP and SCLIP architecture. 

During training we use the AdamW optimizer and the CrossEntropy Loss function applied to the bilinearly upscaled score maps and masks. We observe that the default learning rate from DenseCLIP of 1e-5 works well here. For the final visualization, as seen in fig. 5 (main paper), segmentation predictions are post-processed with PAMR \cite{araslanov2020single} for all visualized models. Note that we do not train for the background class, nor is it queried during the segmentation evaluation for any of the models, specifically to evaluate their ability to segment the parts of an object.

\section{Effect of Shading Cues in Blender-based Rendering}
To be able to isolate and analyse the performance gained from an object-centric dataset like $\text{3DFroMLLM}_\text{GPT-4o}$ we propose to compare the downstream visual performance using  1) 2D renders of prototypes which are naturally shaded using a realistic rendering engine like blender referred to as $\text{3DFroMLLM}^\text{shaded}$ and 2) the same renders only with albedo and silouhette $\text{3DFroMLLM}^\text{albedo}$. This experiment stands as an abridged version of the experiment in section 4.1 in the main paper, and serves solely as a proof-of-concept. The comparison in \cref{tab:shading} shows that even without realistic rendering our method outperforms the baseline, $\text{VCLM}_{\text{GPT-4o}}$. Additionally, shading effects -- which are made possible only through our approach -- provide even better performance and hence are used as part of our method.

\begin{table}
    \centering
    \begin{tabular}{l | S[table-format=2.2]}
        \textbf{Pretraining Data} & {I-100 5NN} \\
        \midrule
        $\text{VCLM}_\text{GPT-4o}$ & 29.68 \\
        $\text{3DFroMLLM}^\text{albedo}$ & 30.86 \\
        $\text{3DFroMLLM}^\text{shaded}$ & \textbf{33.22} \\
    \end{tabular}
    \caption{Effect of shading cues in renders. \emph{Note:} Blender alone allows for guaranteed albedo-only rendering, thus only blender objects are used here.}
    \label{tab:shading}
\end{table}

\section{Prompt Template}
\subsection{Decomposer}

\begin{tcolorbox}[colback=gray!10, colframe=black, sharp corners=southwest, title=\textbf{Prompt Template}, fonttitle=\bfseries]
\small %
Generate a list of key components for a $q$, ensuring that the listed parts are visually distinctive and collectively recognizable as a ``\textbf{\textcolor{PineGreen}q}" when illustrated.

\begin{itemize}
    \item The list should contain \textbf{at least 1} and \textbf{at most \{n\_parts\}} elements.
    \item Additionally, provide a \textbf{count list} indicating the number of times each part typically appears in a ``\textbf{\textcolor{PineGreen}q}".
    \item Output both lists in a \textbf{Python list format}.
\end{itemize}

\textbf{Example Task}  
\textbf{Category:} "Chair"  

\textbf{Output:}  
\begin{lstlisting}[language=Python]
part_list = ['backrest', 'seat', 'legs']
part_counts = [1, 1, 4]
\end{lstlisting}

\textbf{Task:}  
Now, generate the part list for a ``\textbf{\textcolor{PineGreen}q}".

\textbf{Answer:}
\end{tcolorbox}

\subsection{Metricizer}
\begin{tcolorbox}[colback=gray!10, colframe=black, sharp corners=southwest, title=\textbf{Prompt Template}, fonttitle=\bfseries]
\small %
Determine the \textbf{average dimensions} (in \textbf{metric units}) of a \textbf{\textcolor{PineGreen}{q}}, ensuring that the provided values represent the \textbf{overall size} of the object.

\begin{itemize}
    \item The dimensions should describe a \textbf{bounding box} that can fully contain a \textbf{standard instance} of a \textbf{\textcolor{PineGreen}{q}}.
    \item \textbf{Ignore individual part dimensions} and focus only on the total \textbf{length, width, and height} (in that order).
    \item Output the dimensions as a \textbf{Python list}.
\end{itemize}

\textbf{Example Task}  
\textbf{Category:} \textbf{\textcolor{PineGreen}{q}}  

\textbf{Output:}  
\begin{lstlisting}[language=Python]
dimensions = [0.45, 0.45, 1.0] # called 'C' in the main paper
\end{lstlisting}

\textbf{Task:}  
Now, provide the dimensions for a \textbf{\textcolor{PineGreen}{q}} in the same format.

\textbf{Answer:}
\end{tcolorbox}

\vspace{1cm}

\subsection{ArrangementProposer}

\begin{tcolorbox}[colback=gray!10, colframe=black, sharp corners=southwest, title=\textbf{Prompt Template}, fonttitle=\bfseries]
\small %
Generate a \textbf{3D layout JSON representation} of a \textbf{\textcolor{PineGreen}{q}}, using a coordinate convention where:

\begin{itemize}
    \item \textbf{X-axis} extends \textbf{outward} (from back to front).
    \item \textbf{Y-axis} extends \textbf{to the right}.
    \item \textbf{Z-axis} extends \textbf{upward}.
\end{itemize}

\textbf{Key Rules:}
\begin{enumerate}
    \item \textbf{Global Orientation:} Assume the object is \textbf{facing forward} along the X-axis in an \textbf{upright position}. If orientation is ambiguous, use a reasonable assumption.
    \item \textbf{Bounding Box:} The overall dimensions \textbf{\textcolor{PineGreen}{d}} define the object's total size.
    \item \textbf{Part Breakdown:} You will be given a list of \textbf{parts} and their \textbf{instance counts}.
    \item \textbf{Spatial Reasoning:}
    \begin{itemize}
        \item Position each part logically within the given bounding box.
        \item Ensure symmetrical placement where applicable.
        \item Assign a unique label for each instance of a repeated part.
    \end{itemize}
    \item \textbf{Output Format:}
    \begin{itemize}
        \item Represent each part as an \textbf{upright cuboid} with attributes for \textbf{height, width, length, location, and rotation}.
        \item \textbf{Location} is the 3D coordinate of the part's center. This is a 3D tuple/list representing the vector from the origin to the center of the part.
        \item \textbf{Rotation} is given as Euler angles in radians \textbf{[rotation\_x, rotation\_y, rotation\_z]} (e.g., [pi/2,0,0] will rotate a part around the x-axis by 90 degrees).
    \end{itemize}
\end{enumerate}

\textbf{Meta Command:}
\begin{enumerate}
    \item Reason about the global orientation of the object.
    \item For each of the 3 axes, recall the coordinate convention.
    \item Explain what parts of the object will appear in what order when isolating the axis. For this set of parts:
    \begin{itemize}
        \item[(a)] Give a coarse localization of the part in natural language.
        \item[(b)] Reason about the average dimensions of the individual parts.
    \end{itemize}
    \item Given the part average dimensions and coarse localization, generate a JSON object that provides the arrangement of all parts.

    \textbf{Example Task}  
    \textbf{Input:}  
    \textbf{Category:} \textbf{\textcolor{PineGreen}{q}}  
    \textbf{Dimensions:} [0.60, 0.60, 1.00]  
    \textbf{Common Parts:} ["Leg", "Seat", "Backrest"]  
    \textbf{Part Counts:} [4, 1, 1]  
\end{enumerate}

\end{tcolorbox}

\begin{tcolorbox}
\textbf{Step-by-Step Reasoning:}
\\ \dots

\small{\textbf{Final JSON Output:}
\begin{lstlisting}[language=JSON]
{
    "Leg1": {
        "height": 0.45,
        "width": 0.065,
        "length": 0.065,
        "location": [0.04, 0.0325, 0.23],
        "rotation": [0, 0, 0]
    },
    "Leg2": {
        "height": 0.45,
        "width": 0.065,
        "length": 0.065,
        "location": [0.5, 0.0325, 0.23],
        "rotation": [0, 0, 0]
    },
    "Seat": {
        "height": 0.1,
        "width": 0.6,
        "length": 0.6,
        "location": [0.3, 0.3, 0.5],
        "rotation": [0, 0, 0]
    },
    "Backrest": {
        "height": 0.45,
        "width": 0.6,
        "length": 0.1,
        "location": [0.05, 0.3, 0.775],
        "rotation": [0, 0, 0]
    }
}
\end{lstlisting}

\textbf{Your Task:}  
Now, generate the \textbf{3D layout JSON} for a \textbf{\textcolor{PineGreen}{q}}, following the same structured reasoning.  

\textbf{Category:} \textbf{\textcolor{PineGreen}{q}}  \\
\textbf{Dimensions:} \textbf{\textcolor{PineGreen}{$C$}}  \\
\textbf{Common Parts:} \textbf{\textcolor{PineGreen}{${\{l \mid (l,c) \in D\}}$}}  \\
\textbf{Part Counts:} \textbf{\textcolor{PineGreen}{${\{c \mid (l,c) \in D\}}$}} }  \\
\textbf{Answer:}
\end{tcolorbox}

\newpage
\subsection{Proposal2Code}
\begin{tcolorbox}[colback=gray!10, colframe=black, sharp corners=southwest, title=\textb{Prompt Template}, fonttitle=\bfseries]
\small %
You are an expert in \textbf{\textcolor{PineGreen}{l}}, responsible for generating code that constructs a \textbf{3D model} of a \textbf{\textcolor{PineGreen}{q}} based on a provided \textbf{layout proposal} in JSON format.

\textbf{Key Guidelines:}
\begin{itemize}
    \item Each part is represented as a \textbf{cuboid primitive}.
    \item The JSON object provides parameters:
    \begin{itemize}
        \item \textbf{Height, Width, Length} (size of the cuboid).
        \item \textbf{Location}: A \textbf{3D coordinate} \textbf{[X, Y, Z]} specifying the center position.
        \item \textbf{Rotation}: A \textbf{3D tuple [θ\_x, θ\_y, θ\_z]} representing \textbf{Euler angles} in \textbf{radians}.
    \end{itemize}
    \item \textbf{Coordinate Convention:}
    \begin{itemize}
        \item \textbf{X-axis (Length)}: Points \textbf{outward} (back to front).
        \item \textbf{Y-axis (Width)}: Points \textbf{to the right}.
        \item \textbf{Z-axis (Height)}: Points \textbf{upward}.
    \end{itemize}
    \item \textbf{Rotation Handling:}
    \begin{itemize}
        \item Only apply rotations \textbf{when necessary}.
        \item Prefer adjusting \textbf{scale and location} before modifying orientation.
    \end{itemize}
\end{itemize}

---

\textbf{Example Task}  
\textbf{Input:}  
\textbf{Category:} \textbf{\textcolor{PineGreen}{q}}  \\
\textbf{Layout Proposal:}  
\begin{lstlisting}[language=JSON]
{
    "Seat": {
        "height": 0.1,
        "width": 0.6,
        "length": 0.6,
        "location": [0.3, 0.3, 0.5],
        "rotation": [0, 0, 0]
    },
    "Backrest": {
        "height": 0.45,
        "width": 0.6,
        "length": 0.1,
        "location": [0.05, 0.3, 0.775],
        "rotation": [0, 0, 0]
    }
    "Leg1": {
        ...
    },
    "Leg2": {
        ... <redacted for brevity>
    },
    "Leg3": {
        ...
    },
    "Leg4": {
        ...
    },
}
\end{lstlisting}
\end{tcolorbox}

\begin{tcolorbox}
\small %

---

\textbf{Expected Output:}  
...

---

\textbf{Your Task:}  
Now, generate the \textbf{Blender Python script} for a \textbf{\textcolor{PineGreen}{q}}, following the structured approach above.  

\textbf{Layout Proposal:}  
\textbf{\textcolor{PineGreen}{A}}

\textbf{Answer:}
\end{tcolorbox}

\vspace{1cm}
\subsection{Code Refiner}
\begin{tcolorbox}[colback=gray!10, colframe=black, sharp corners=southwest, title=\textbf{Prompt Template}, fonttitle=\bfseries]
\small %
The provided code \textbf{does not accurately represent} a \textbf{\textcolor{PineGreen}{q}}. Your task is to \textbf{improve the representation} while maintaining the overall structure as much as possible.

\textbf{Allowed Modifications:}
\begin{itemize}
    \item \checkmark \textbf{Enhance Accuracy:} Adjust the \textbf{position, orientation, and proportions} of parts.
    \item \checkmark \textbf{Modify Geometry:}
    \begin{itemize}
        \item \textbf{Add} missing parts or elements.
        \item \textbf{Rearrange} existing parts.
        \item \textbf{Redraw} parts using more appropriate \textbf{primitive types} (e.g., using cylinders instead of cubes for legs).
    \end{itemize}
    \item \checkmark \textbf{Adjust Materials:} Assign \textbf{new materials} to enhance realism or clarity.
    \item \checkmark \textbf{Transform Shapes:} Modify, reshape, or scale elements to better match the intended design.
    \item \checkmark \textbf{Remove Parts (If Necessary):} Only remove components if they \textbf{inaccurately} contribute to the model.
\end{itemize}

\textbf{Constraints:}
\begin{itemize}
    \item \textbf{Preserve Structural Integrity:} Ensure changes \textbf{do not break} the overall structure.
    \item \textbf{Maintain Proper Orientation \& Positioning:} Carefully consider \textbf{axes alignment} and \textbf{symmetry} in the model.
\end{itemize}

---

\textbf{Old Code:}  
\textcolor{PineGreen}{$P_{coarse}$}

\textless or \textgreater

\textbf{Edit Recommendations}  
\textcolor{PineGreen}{$E_i$}

---

\textbf{Updated Code:}  
\end{tcolorbox}

\newpage
\subsection{Identifier}

\begin{tcolorbox}[colback=gray!10, colframe=black, sharp corners=southwest, title=\textbf{Prompt Template}, fonttitle=\bfseries]
\small %
You are given \textbf{rendered images} (\textcolor{PineGreen}{$R_0$}) of a \textbf{prototypical object} modeled in \texttt{PMA}. Your task is to \textbf{identify the object} based on its \textbf{general structure, arrangement, and proportions}.

\textbf{Key Considerations:}
\begin{itemize}
    \item \checkmark \textbf{Multiple Views:} Each image shows the object from \textbf{\textcolor{PineGreen}{s}} different angles, with the object \textbf{centered} in the frame.
    \item \checkmark \textbf{Focus on Geometry:} The model consists of \textbf{primitive shapes} with \textbf{missing fine details}—ignore material properties.
    \item \checkmark \textbf{Prioritize Structural Features:} Base your predictions on the \textbf{shape, geometry, and arrangement} of the object's parts.
    \item \checkmark \textbf{Provide Multiple Guesses:} List up to \textbf{\textcolor{PineGreen}{p}} reasonable predictions.
\end{itemize}

\textbf{Expected Output Format:}
\begin{itemize}
    \item The output must be a \textbf{valid Python list} containing \textbf{your top predictions} for the object.
    \item Ensure predictions are \textbf{concise and relevant}.
\end{itemize}

---

\textbf{Example Output:}  
predictions = [`Chair', `Laptop', `Street Lamp']

---

\textbf{Your Output:}  
\begin{lstlisting}[language=Python]
# Your predictions here
\end{lstlisting}

\end{tcolorbox}

\subsection{Edit Recommender}

\begin{tcolorbox}[colback=gray!10, colframe=black, sharp corners=southwest, title=\textbf{Prompt Template}, fonttitle=\bfseries]
\small %
You previously predicted that the given renders depicted \textbf{one of the following objects: \textcolor{PineGreen}{$\hat{\textbf{q}}_0$}}, while the actual object was a \textbf{\textcolor{PineGreen}{q}}.

Your task is to:
\begin{itemize}
    \item ✅ \textbf{Identify misleading visual features} that contributed to the incorrect prediction.
    \item ✅ \textbf{Provide precise refinement instructions} to improve the model.
    \item ✅ \textbf{Format your response} as a \textbf{list of dictionary items}, each describing a feature to adjust.
\end{itemize}

\textbf{Guidelines for Refinements:}
\begin{itemize}
    \item Each refinement must fall under one or more of these \textbf{types}: \textbf{[add, remove, move, stretch, scale]}.
    \item Each refinement must clearly refer to the \textbf{problematic part or visual feature}.
    \item Instructions should be written to assist a \textbf{3D artist} in making corrections.
    \item You may provide \textbf{up to 5 refinement instructions}.
\end{itemize}

---
\end{tcolorbox}

\begin{tcolorbox}

\textbf{Expected Output Format:}
example = [
    {
        ...
    },
    ...
]

---

\textbf{Example Output (Chair Misidentified as a Table):}  
$E_0$ = [
    {
        ...
    }
]

---

\textbf{Your Output:}  
\begin{lstlisting}[language=Python]
# Your refinements here
\end{lstlisting}    
\end{tcolorbox}

\subsection{Prompt Template for Non-Agentic Baseline}
\begin{tcolorbox}[colback=gray!10, colframe=black, sharp corners=southwest, title=\textbf{Prompt Template}, fonttitle=\bfseries]
\small %
Write a code that draws \textbf{\textcolor{PineGreen}{q}}
\end{tcolorbox}

\section{GPT-4o Setting}
For our \texttt{MLLM-Engine} we chose the temperature value 0.2 to ensure predictable and typical results for prototype arrangement and design. All datasets and evaluations were done with the snapshot version: 
\texttt{gpt-4o-2024-08-06}

\section{Textured Mixup Dataset}
For texturing our dataset we use the same strategy as described in \cite{vclm}. All dataset have a size 1.3M. Each image is generated by mixup up two samples chosen at a 50\% probability either from the contending dataset ($\text{3DFroMLLM}_\text{GPT-4o}, \text{VCLM}_\text{GPT-4o}$) or Shaders21k \cite{baradad2022procedural}. All other training protocols are identical to the ones in the main text.

\section{Limitations}
Hallucinations \cite{huang2025survey} commonly plague generative models especially in large planning tasks. Regardless of compilation fail-safes and API retries, run-time errors may appear. Language-specific fail-safes allow for procedural code to run producing incomplete 3D prototypes or sometimes empty renders. 

Furthermore, deciding what is and what isn't a part is non-trivial due to inherent difficulty of constructing taxonomies \cite{chen2020constructing}. Knowing the depth one has to reach in an object taxonomy tree, such that a visually compelling output can be generated, is an even more arduous task. This is why in our final renders, prototypes can have parts logically arranged but individual parts may not be aligned to the physical world (i.e. parts are floating, not attached via other connective sub-parts). In future works, we hope to dig deeper into how part-taxonomies can be reliably extracted to generate even higher fidelity and physically-aligned, parametric 3D models.

\vspace{50pt}

\section{Sparse Evaluation Pseudo Code}
\begin{algorithm}
\caption{Main Execution Flow for Sparse Geometric Evaluation}
\begin{algorithmic}[0]

\Function{sparse\_evaluation}{prototype, targets}
    \For{t in targets}
        \State mapping $\gets$ \Call{map}{prototype.parts, t.parts}
        \State \Call{umeyama}{prototype, t, mapping} $\gets$ transform
        \State alignedProto $\gets$ transform @ prototype
        \State \textit{residuals} $\gets$ \textit{residuals} $\cup$ \Call{computeResidual}{target, alignedProto}
    \EndFor
    \State \Return \Call{computeNNMetric}{residuals, 5}    
\EndFunction

\end{algorithmic}
\end{algorithm}

\bibliography{aaai2026}